\pdfoutput=1

\documentclass[11pt]{article}

\usepackage[]{emnlp2021}

\usepackage{times}
\usepackage{latexsym}

\usepackage[T1]{fontenc}

\usepackage[utf8]{inputenc}

\usepackage{microtype}

\usepackage{graphicx}
\usepackage{svg}
\usepackage{subcaption}
\usepackage{enumitem}
\usepackage{booktabs,subcaption,amsmath}
\usepackage{multirow}
\usepackage{makecell}
\usepackage{hyperref}
\usepackage{caption}
\usepackage{cleveref}
\usepackage{textcomp}
\usepackage{arydshln}

\crefname{section}{\S}{\S\S}
\Crefname{section}{\S}{\S\S}
\crefname{table}{Table.}{}
\crefname{figure}{Figure}{}
\crefname{algorithm}{Algorithm}{}
\crefname{equation}{equation}{}
\crefname{appendix}{Appendix}{}
\crefformat{section}{\S#2#1#3}  

\graphicspath{ {./images/} }
\newcommand{\err}[1]{\textsubscript{~$\pm$#1}}

\title{Translate \& Fill: Improving Zero-Shot Multilingual Semantic Parsing with Synthetic Data}

\author{Massimo Nicosia, Zhongdi Qu, Yasemin Altun\\\\
  Google Research \\
  \texttt{\{massimon,dqu,altun\}@google.com} \\}

\begin{document}
\maketitle
\begin{abstract}
While multilingual pretrained language models (LMs) fine-tuned on a single language have shown substantial cross-lingual task transfer capabilities, there is still a wide performance gap in semantic parsing tasks when target language supervision is available. In this paper, we propose a novel Translate-and-Fill (TaF) method to produce silver training data for a multilingual semantic parser. This method simplifies the popular Translate-Align-Project (TAP) pipeline and consists of a sequence-to-sequence filler model that constructs a full parse conditioned on an utterance and a view of the same parse. Our filler is trained on English data only but can accurately complete instances in other languages (i.e., translations of the English training utterances), in a zero-shot fashion. Experimental results on three multilingual semantic parsing datasets show that data augmentation with TaF reaches accuracies competitive with similar systems which rely on traditional alignment techniques.
\end{abstract}

\section{Introduction}

Semantic parsing is a core task in virtual assistants, popular applications that require accurate natural language understanding (NLU). User utterances are parsed into a structured representation made of intents and slots that is interpreted to initiate an action on the user device. For example, the sentence ``\emph{set an 8 am alarm}'' could lead to the following interpretation -- \emph{Create\_alarm(time=``8 am'')} -- and result in an alarm being created.

As in many NLP tasks, numerous English parsing datasets are available and well studied \cite{price-1990-evaluation, banarescu-etal-2013-abstract, dialog-state-tracking, fan-etal-2017-transfer, gupta-etal-2018-semantic,goo-etal-2018-slot, qin-etal-2019-stack, 10.1145/3366423.3380064}. Supporting new domains and schemas requires a sizeable data collection effort and while English is receiving the most attention, it is also important to extend NLU to other languages in order to provide users consistent experiences across languages. 
Multilingual pretrained language models (LMs) are an excellent starting point for enabling cross-lingual transfer in a parser but they are no substitute for using high quality, albeit costly to annotate, training data in the target languages.
Without such data, we can translate the available annotated examples to other languages and slot annotations can be transferred \cite{Yarowsky2001, Shah10synergy:a}. Traditionally, annotation transfer requires (i) token alignment models \cite{brown-etal-1993-mathematics}, which may have been trained on text tokenized differently from the annotated training data, and (ii) label projection logic that can be complex, especially if it includes heuristics for fixing systematic alignment errors, or if nested structures need to be mapped.

In this work, we propose an alternative approach to the classical Translate-Align-Project (TAP) pipeline: we leverage multilingual pretrained representations and a sequence-to-sequence (seq2seq) model to directly generate the parse of translated examples in a zero-shot fashion. Our model is trained on English data only and it is able to reconstruct the full parse while having access to the English utterance and to a signature (or view) of the full parse. At inference, we substitute the English utterance with its translation and our model, pulling content from the latter, is able to construct a high quality silver parse. The main contributions of this paper can be summarized as follows:

\begin{itemize}
    \item We propose a novel approach, Translate-and-Fill (TaF), for generating synthetic data to train multilingual semantic parsers that is robust to tokenization, is inherently generative and makes use of the intent and slot schema to potentially learn label-specific alignment rules.
    TaF replaces the alignment and projection modules of the TAP approach with a learned component that generates full parses of examples translated from English, removing the need of aligners.
    \item We analyze the zero-shot capabilities of TaF in terms of quality of the silver parses.
    \item We evaluate the impact of the synthetic data generated with our approach on three multilingual semantic parsing datasets,
    showing that data augmentation with TaF on multilingual pretrained seq2seq models sets new state-of-the-art (SOTA) results in multiple scenarios and in some cases, closes the gap with respect to full in-language supervision.
\end{itemize}

\section{Related Work}

Our work is closely related to two research areas: (i) multilingual representations and models, and (ii) annotation projection methods.

Cross-lingual transfer has been studied in several structured prediction tasks such as part-of-speech tagging \cite{Yarowsky2001, Tackstrom2013, Plank2018, Kann_Lacroix_Sogaard_2020}, named entity recognition \cite{Zirikly2015,Tsai2016,Xie2018} and dependency parsing \cite{guo-etal-2015-cross, ahmad-etal-2019-cross, zhang-etal-2019-cross}.

One way to achieve cross-lingual transfer is by adopting multilingual representations and models pretrained on a large amount of text in different languages. This way, similar languages with overlapping vocabularies at word or subword level can benefit from information sharing. These models can encode the input using words \cite{Word2Vec, pennington-etal-2014-glove}, characters or subwords \cite{sennrich-etal-2016-neural, kudo-richardson-2018-sentencepiece, 45610, DBLP:journals/corr/abs-2103-06874}. With the latter, interesting zero-shot performance (i.e., training on a language and evaluating on a different target language) can be achieved, especially between similar languages \cite{lauscher-etal-2020-zero}.  

Multilingual representations can be obtained from encoders pretrained on multilingual corpora with tasks such as masked language modeling (MLM), or trained on supervised tasks such as neural machine translation (NMT) \cite{Eriguchi2018, Yu2018, Singla2018, Siddhant_Johnson_Tsai_Ari_Riesa_Bapna_Firat_Raman_2020}. After the success of fill-in-the-blank-style denoising objectives and BERT/mBERT \cite{Devlin2019}, other multilingual encoders achieved a similar level of popularity. These models include XLM \cite{LampleC2019}, XLM-R \cite{Conneau20XLMR} and a recent multilingual version of T5 \cite{JMLR:v21:20-074}, named mT5 \cite{xue2021mt5}. T5 based models differ from the others by their seq2seq architecture where both the encoder and the decoder are pretrained with the MLM task. In this work, we leverage the multilinguality and the generative capabilities of mT5 to produce interpretations and create synthetic internationalization (i18n) data for semantic parsing.

\begin{figure*}
  \includegraphics[width=\textwidth]{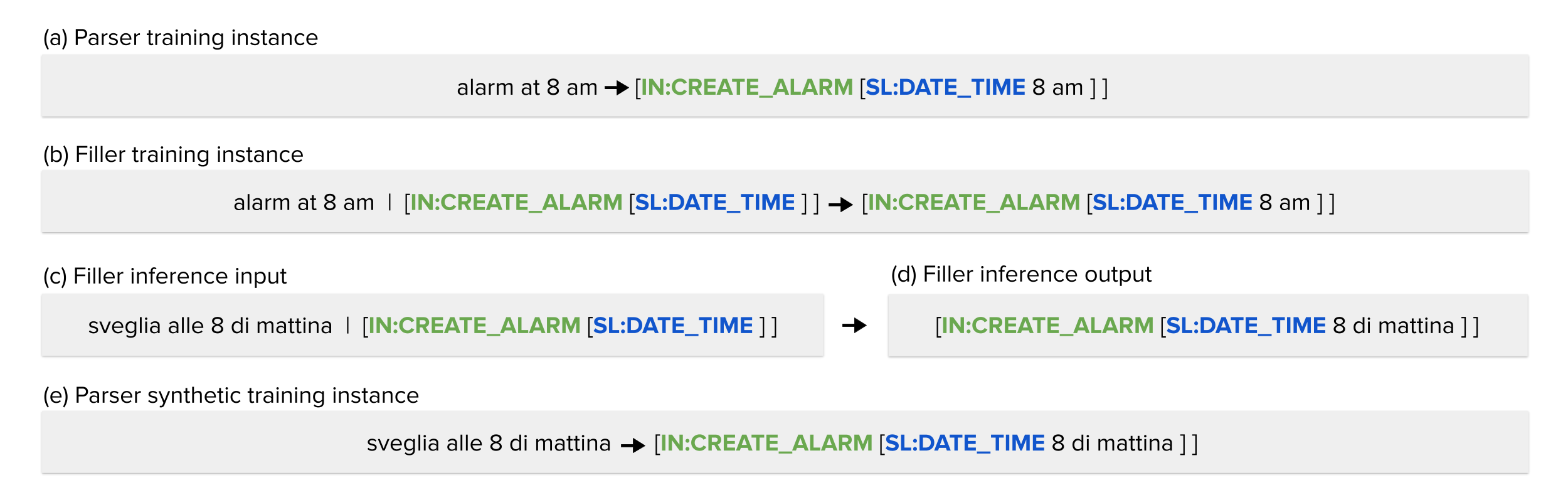}
  \footnotesize
  \caption{Example instances for training the semantic \textbf{parser} (a) and the \textbf{filler} (b). The filler is trained to produce a full parse from the concatenation of an English utterance and the corresponding parse signature (b). At inference, we replace the English utterance with its (Italian in this case) translation (c), and obtain a silver parse where the slots contain text from the translation (d). The latter is used to assemble a synthetic training instance (e) for a multilingual semantic parser.}
  \label{instances}
\end{figure*}

A second way to improve cross-lingual transfer is data augmentation. Typically, annotated data is available in at least one language, and more often than not, this is a high-resource language such as English. NMT is a strong data augmentation baseline, as shown in recent cross-lingual evaluation benchmarks \cite{pmlr-v119-hu20b, ladhak-etal-2020-wikilingua}. NMT can be used to translate training examples from a source to a target language (translate-train), creating training data in the target language. Otherwise, it can be used to translate the test data to the language of the trained model (translate-test).

While translating works quite well for classification tasks where the label is at instance level, for sequence tagging or parsing tasks the reality is more challenging since the labels are at token level and they have to be transferred from the tokens of the original text to the tokens of its translation.

Prior work relies on word aligners to establish a match between the tokens of source and translated text, and to transfer the labels \cite{Ni2017, Jain2019, daza-frank-2020-x, fei-etal-2020-cross}. Alignment methods include unsupervised word alignment \cite{brown-etal-1993-mathematics, vogel-etal-1996-hmm, och-ney-2000-improved, och03:asc}, the use of attention weights from NMT models \cite{Schuster2019, chen-etal-2020-accurate, zenkel-etal-2020-end} or computing the similarity between word embeddings \cite{jalili-sabet-etal-2020-simalign, dou-neubig-2021-word}.

In this work, we propose an alternative and novel label projection method that leverages the signatures of available parses for internationalization, in the spirit of sketch or template decoding \cite{dong-lapata-2018-coarse, zhang-etal-2019-adansp, wiseman-etal-2018-learning}). 
Our method avoids alignment models altogether and leverages multilingual representations and instance labels to generate high quality silver data that can be finally used to train accurate multilingual semantic parsers. In addition and differently from NMT attention-based aligners, our method does not access the internals of neural translation models and therefore has a wider applicability.

\section{Translate-and-Fill (TaF)}

We address the problem of the i18n of semantic parsers when (i) English training data is available and (ii) high quality and cost-effective training data in other languages is needed.
We translate English data to a target language using NMT. This leaves us with the problem of mapping original slot annotations to translations. Our solution is a novel method that we call Translate-and-Fill (TaF), which replaces the align and project modules of the popular Translate-Align-Project (TAP) pipeline while leveraging multilingual pretraining.
In our approach, we use two seq2seq models trained differently: one is the usual \textbf{semantic parser} and the other is what we call the \textbf{filler}.

Figure \ref{instances} shows the example instances used to train the semantic parser and our filler, and then to run inference with the latter. The example parse has a \texttt{CREATE\_ALARM} intent (\texttt{IN:}) and a single \texttt{DATE\_TIME} slot (\texttt{SL:}). We transform a training instance for the semantic parser that maps an utterance to its parse (a) into a training instance for the filler. A filler training instance (b) maps the English utterance concatenated with its parse signature to a full parse (target from a). To obtain the parse signature, we simply remove all the slot values from the parse. The filler must then reproduce the input signature while filling the signature slots with words from the input utterance. 

We leverage pretrained multilingual seq2seq models (in particular mT5) to train the filler model with only English filler instances.
A trained filler can be used to obtain labeled semantic parsing data in other languages, thanks to the cross-lingual transfer capabilities of the pretrained seq2seq model, as well as the slot-filling capabilities gathered from the English training filler instances.
We construct an inference example for the filler from the same examples used to train it (b) by simply replacing the English utterance in the input part with its corresponding translation (c). The filler will now reproduce the input signature but fill the slots using words from the translation (d).

Finally, we create a synthetic i18n instance for training a parser for the target language. The synthetic instance maps the translated utterance to the parse produced by the filler at inference (e).

Similar to TAP, our basic assumption is that the parse structure of a translated sentence does not change.
The proposed approach (i) can be applied to any language supported by NMT and by the pretrained seq2seq model; (ii) can handle nested interpretations naturally thanks to the seq2seq formulation; and (iii) since it has access to the interpretation, it can learn label specific projection strategies as opposed to the handcrafted TAP projection rules.

\section{Experimental Setup}

\subsection{Datasets}

We experiment with three multilingual task-oriented semantic parsing datasets.
\\\\
\textbf{MTOP} \cite{li-etal-2021-mtop} is an almost parallel dataset covering 6 languages and 11 domains. Each utterance has associated intent and slots, but also comes with a decoupled compositional representation similar to the parses in Figure \ref{instances}. Compositional instances will have nested intents. The seq2seq nature of our model lets us handle such cases without any specialized component. For our experiments, we use the provided train/validation/test splits and focus on predicting the decoupled representations. 
\\\\
\textbf{Multilingual ATIS} \cite{mulitatis2018} is a dataset for the travel-planning domain that extends the popular ATIS dataset \cite{price-1990-evaluation} to two other languages: Hindi and Turkish. Differently from MTOP, there are no nested intents and therefore just flat span annotations. 
\\\\
\textbf{MultiATIS++} \cite{xu-etal-2020-end} adds six new languages to Multilingual ATIS bringing the number of non-English languages to 8. For both Multilingual ATIS and MultiATIS++, we create an MTOP-style interpretation by converting the BIO-tagged sentences into an intent/slot structure (as in Figure \ref{instances}). For both datasets we use the train/validation/test split ratios reported in \newcite{xu-etal-2020-end}.

\subsection{Models}

Our parser and filler are trained using mT5 \cite{xue2021mt5}, a multilingual version of the text-to-text T5 model \cite{JMLR:v21:20-074}, pretrained on the mC4 corpus and 101 languages. We experiment with two mT5 versions, \texttt{large} and \texttt{xxl} in different settings. In the \textit{gold data} setting, we train multilingual parsers with all the available training data. In the \textit{zero-shot} setting, we train our models on English data only. In the \textit{+TaF} setting, we train our models on English gold data and on the synthetic data produced by our filler for all the other languages.
We do not do any hyper-parameter tuning and use a batch size of 512 and a constant 0.001 learning rate. We train all our models for 3k steps saving checkpoints every 200 steps.
The parser produces structured interpretations and we run Unicode normalization on the tokens.
The filler is an mT5-xxl model trained for 400 steps since its output does not significantly change after that. We then run inference on the << translation | signature >> inputs and generate synthetic training data for the parser. Apart from discarding a negligible number of outputs that cannot be parsed into a tree, we do not apply any additional quality filter.
According to our experience, this is an advantage w.r.t. alignment-based methods that require complex filtering to suppress systematic alignment errors and improve synthetic data quality.

\subsection{Translation and Postprocessing}

We translate the English utterances to different target languages and tokenize them with in-house translation and tokenization systems. The datasets used in our experiments come with tokenized gold data but no tokenized translations. In the MTOP paper an in-house tokenizer is used, while the other dataset papers do not contain details about tokenization. This is a common issue, as also reported in \newcite{DBLP:journals/corr/abs-2101-08890}. This implies a tokenization mismatch between our synthetic data and the synthetic data used in the original dataset papers. To quantify this, we compare our tokenization of MTOP utterances with gold tokenization. Table \ref{tab:tokenization} shows that we can reasonably match the original tokenization for all languages except for Thai. In the synthetic data setting, this could potentially disadvantage our results due to the noise introduced by the dissimilar tokenization.
In one experiment, we do not tokenize the translations to test the quality of the final synthetic data.
 
In Multilingual ATIS and MultiATIS++, Spanish and Turkish eval data is all lowercase and we lowercase our translations too. In addition, Turkish data does not contain special characters, so we replace the latter in the translations according to the following mapping: \texttt{ğĞıİöÖüÜşŞçÇ $\Rightarrow{}$gGiIoOuUsScC}.

\begin{table}
\setlength{\tabcolsep}{0.6em}
\renewcommand{\arraystretch}{1.2}
\scriptsize
\centering
\begin{tabular}{lcccccc}
\toprule
{\bf Language } & en & de & es & fr & hi & th \\
{\bf Match \% } & {93.50} & {93.75} & {96.39} & {94.61} & {98.35} & {42.08} \\
\bottomrule
\end{tabular}
\caption{\% of MTOP training instances where our tokenization matches the original MTOP tokenization.}
\label{tab:tokenization}
\end{table}

\subsection{Evaluation}

For MTOP, we report Exact Match (EM) accuracy as in \newcite{li-etal-2021-mtop}.
For Multilingual ATIS and MultiATIS++, we report EM accuracy, Intent Accuracy and Slot F1 (micro) computed with the \texttt{seqeval} toolkit \cite{seqeval}.
Since we predict structured interpretations, we reconvert our outputs to a sequence of BIO-tagged tokens before computing Slot F1. We first map slots which can be unambiguously identified in the input utterance by full or partial string matching.
The remaining slots are aligned using the Needleman-Wunsch alignment algorithm \cite{Needleman1970AGM}, a strategy shown to be robust to small generation errors \cite{tanl}.
In the \textit{Avg} columns of the tables, we report the evaluation metrics averaged over the non-English languages.

For the \textit{gold} data setting, we do model selection by computing the best average (across non-en languages) EM on the dev set. For the \textit{zero-shot} and \textit{+TaF} settings we compute metrics using the last checkpoint, assuming the unavailability of a development set. To keep the amount of compute required for running the experiments reasonable, all our numbers are averaged over three runs, and we report standard deviation.

\subsection{Translate-and-Align (TAP) Baseline}
We also experiment with synthetic data produced via TAP, aligning tokens with an implementation of the IBM Model \cite{brown-etal-1993-mathematics} and HMM \cite{vogel-etal-1996-hmm}. To achieve the best alignment quality, we tokenize both the English input and the translations with our in-house tokenizer (also used to train the aligner), and discard examples for which our tokenization of the English utterance differs from the original. We apply heuristic filters to the synthetic data, discarding examples where a span is split into non-consecutive tokens in the target, and examples where the target has a set of slots different from the source.

Two significant sources of error in TAP data are prepositions and determiners. When those are introduced in a translation, they are often aligned to the adjacent nouns in the original English utterance and are therefore included in the nouns' slots. Take an example from the MTOP dataset, ``Play some Elvis for me''. Its French translation is ``Jouez à Elvis pour moi'', where ``à'' is a preposition with no direct correspondence in the English utterance. As a result, our aligner maps it to ``Elvis'', and the value for the slot \texttt{MUSIC\_ARTIST\_NAME} becomes ``à Elvis'', instead of ``Elvis''. To mitigate this problem, we run the translated utterances through an in-house parts-of-speech tagger and exclude prepositions and determiners from the slots when they appear at slot boundaries (except for the slot \texttt{DATE\_TIME}, for which prepositions and determiners are generally kept as a part of the slot values in the MTOP dataset). The POS tagger also performs tokenization and we discard examples for which the POS tokenization differs from the aligner tokenization so that the data left have both high-quality alignments and POS tags.

We also observed that the aligner performs poorly around punctuations that are introduced in the target utterances to function as word connectors. Take an example from MTOP, ``will there be fog in the morning". Its French translation is ``y aura-t-il du brouillard le matin", where ``il" translates to ``it" and serves the same function as ``it" in English sentences about the weather such as ``it is raining". Our aligner maps both the second ``-" and ``il" to ``fog", and as a result the value for the slot 
\texttt{WEATHER\_ATTRIBUTE} becomes ``- il du brouillard", instead of just ``brouillard". To obtain high-quality synthetic data without these issues, we have experimented with training using only the part of data where our tokenizer does simple white-space tokenization on the target utterances. These data points, which do not contain punctuations as individual tokens, are easier to align and ultimately leads to better synthetic data.

The fraction of examples discarded during the TaF filtering stage ranges between 0.01\%-0.4\% for both MTOP and MultiATIS++. For TAP, significantly more filtering was required: for MTOP, 33.1\% of examples were filtered because the aligner tokenizes the source queries differently from the dataset tokenizer, 30.4\% because target queries cannot be simply tokenized by white-space, 0.8\% due to span splitting, and 3.1\% because projected labels have a different set of slots from the original signature; for MultiATIS++, 10.0\% were filtered because the aligner tokenization differs from the provided source tokens, 32.9\% because our tokenizer and the aligner tokenize the translations differently, 0.8\% because of span splitting, and 5.8\% because projected labels have a different set of slots from the original signature.

\section{Results and Discussion}

%
%

\begin{table*}
\begin{center}
\scriptsize
\setlength{\tabcolsep}{1.2em}
\renewcommand{\arraystretch}{1.3}
\begin{tabular}[b]{p{4cm}lllllll}
\toprule
\multirow{1}{*}{\bf MTOP} & \bf en & \bf es & \bf fr & \bf de & \bf hi & \bf th & \bf{Avg}(5 langs) \\
\midrule
\multicolumn{8}{l}{\textit{Multilingual models (trained an all data from all languages)}} \\
\midrule
XLM-R  & 83.6 & 79.8 & 78 & 74 & 74 & 73.4 & 75.8\\
mT5-large  & 83.8\err{0.2} & 76.9\err{0.1} & 75.2\err{0.2} & 72.8\err{0.3} & 73.2\err{0.4} & 73.3\err{0.2} & 74.3\err{0.2} \\
mT5-xxl  & 86.0\err{0.4} & 79.3\err{0.6} & 77.5\err{0.5} & 75.5\err{0.9} & 75.7\err{0.3} & 75.1\err{0.3} & 76.6\err{0.5} \\
\midrule
\multicolumn{8}{l}{\textit{Zero-shot models (trained on English only)}} \\
\midrule
XLM-R  & N/A & 50.3 & 43.9 & 42.3 & 30.9 & 26.7 & 38.8 \\
mT5-large  & 83.2\err{0.2} & 40.0\err{0.7} & 41.1\err{1.8} & 36.2\err{1.5} & 16.5\err{3.3} & 23.0\err{2.1} & 31.3\err{1.8} \\
mT5-xxl  & 86.7\err{0.1} & 62.4\err{2.1} & 63.7\err{1.3} & 57.1\err{1.2} & 43.3\err{0.2} & 49.2\err{0.8} & 55.1\err{1.0} \\
\midrule
\multicolumn{8}{l}{\textit{Augmented data models}} \\
\midrule
XLM-R + TAP  & N/A & \textbf{71.9} & 70.3 & 62.4 & \textbf{63} & 60 & 65.5 \\
mT5-large + TaF  & 83.5\err{0.6} & 69.6\err{0.7} & 71.1\err{0.6} & 70.5\err{0.4} & 58.1\err{1.1} & 57.5\err{0.5} & 65.4\err{0.6} \\
mT5-xxl + TaF & 85.9\err{0.1} & 71.5\err{0.2} & 74.0\err{1.1} & \textbf{72.4}\err{0.2} & 61.9\err{0.4} & 60.2\err{0.3} & 68.0\err{0.1} \\
mT5-xxl + TaF, untokenized  & 85.9\err{0.2} & 71.5\err{0.1} & \textbf{74.6}\err{0.2} & 71.9\err{0.1} & 61.5\err{0.4} & \textbf{62.2}\err{0.4} & \textbf{68.3}\err{0.1} \\
mT5-xxl + TAP  & 86.2\err{0.1} & 69.3\err{0.4} & 71.5\err{0.3} & 62.1\err{0.3} & 57.8\err{0.3} & 58.2\err{0.9} & 63.8\err{0.4} \\
\bottomrule
\end{tabular}
\caption{Exact Match (EM) accuracies on the MTOP dataset. XLM-R results are from \newcite{li-etal-2021-mtop}. In bold, we mark best performances in the data augmentation scenario.}
\label{tab:mtop}
\end{center}
\end{table*}

\begin{table*}
\begin{center}
\scriptsize
\setlength{\tabcolsep}{1.1em}
\renewcommand{\arraystretch}{1.3}
\begin{tabular}[b]{p{4cm}cccccc}
\toprule
\multirow{1}{*}{\bf MTOP} & \bf es & \bf fr & \bf de & \bf hi & \bf th & \bf{Avg}\\
\midrule
mT5-xxl (zero-shot) & 62.4 & 63.7 & 57.1 & 43.3 & 49.2 & 55.1 \\
mT5-xxl + TAP & 54.2 & 55.8 & 57.4 & 55.3 & 39.8 & 52.5 \\
\phantom{--} + POS-based postprocessing & 68.5 & 67.2 & 62.2 & 59.6 & 46.0 & 60.7 \\
\phantom{--} + white-space tokenization & 69.3 & 71.5 & 62.1 & 57.8 & 58.2 & 63.8 \\
\bottomrule
\end{tabular}
\caption{Exact Match (EM) on the MTOP dataset with different TAP configurations.}
\label{tab:mtop-lc}
\end{center}
\end{table*}

%
%

\begin{table*}
\setlength{\tabcolsep}{0.20em}
\renewcommand{\arraystretch}{1.3}
\scriptsize
\begin{center}
\begin{tabular}[b]{p{3cm}llllllllll}
\toprule
\multirow{1}{*}{\bf MultiAtis++} & \bf en & \bf es & \bf de & \bf zh & \bf ja & \bf pt & \bf fr & \bf hi & \bf tr & \bf Avg(8 langs) \\
\midrule
\multicolumn{7}{l}{\textit{Multilingual Intent Accuracy}} \\
\midrule
mBERT & 97.20 & 96.77 & 96.86 & 95.54 & 96.44 & 96.48 & 97.24 & 92.70 & 92.2 & 95.44 \\
mT5-xxl & 97.84\err{0.13} & 97.57\err{0.17} & 97.16\err{0.17} & 97.13\err{0.26} & 97.50\err{0.17} & 97.72\err{0.26} & 97.98\err{0.22} & 95.97\err{0.51} & 94.87\err{0.40} & 96.99\err{0.27} \\
\midrule
\multicolumn{7}{l}{\textit{Multilingual Slot F1}} \\
\midrule
mBERT & 95.90 & 87.95 & 95.00 & 93.67 & 92.04 & 91.96 & 90.39 & 86.73 & 86.04 & 91.02 \\
mT5-xxl & 96.29\err{0.04} & 89.31\err{0.39} & 95.48\err{0.16} & 94.59\err{0.21} & 93.54\err{0.03} & 93.00\err{0.27} & 90.12\err{0.11} & 89.83\err{0.25} & 87.88\err{0.20} & 91.72\err{0.20} \\
\midrule
\multicolumn{7}{l}{\textit{Zero-Shot and Augmented Intent Accuracy}} \\
\midrule
mBERT & N/A & 96.35 & 95.27 & 86.27 & 79.42 & 94.96 & 95.92 & 80.96 & 69.59 & 87.34 \\
mBERT + fastalign & N/A & 97.02 & 96.77 & 96.10 & 88.82 & 96.55 & 96.89 & 93.12 & 93.77 & 94.88 \\
mBERT + softalign & N/A & 97.20 & 96.66 & 95.99 & 88.33 & 96.78 & 97.49 & 92.81 & 93.71 & 94.87 \\
\hdashline
mT5-xxl & 97.87\err{0.11} & 96.90\err{0.34} & 93.06\err{1.62} & 92.53\err{0.55} & 89.18\err{0.64} & 96.75\err{0.22} & 96.83\err{0.42} & 92.46\err{0.32} & 86.67\err{1.07} & 93.05\err{0.47} \\
mT5-xxl + TaF  & 97.65\err{0.11} & 97.65\err{0.22} & 96.79\err{0.13} & 96.75\err{0.11} & \textbf{95.41}\err{0.19} & \textbf{97.61}\err{0.17} & \textbf{97.61}\err{0.17} & \textbf{96.53}\err{0.11} & \textbf{95.06}\err{0.21} & \textbf{96.68}\err{0.12} \\
mT5-xxl + TAP & 97.76\err{0.11} & \textbf{97.69}\err{0.06} & \textbf{97.76}\err{0.11} & \textbf{97.72}\err{0.26} & 94.66\err{0.53} & 96.79\err{0.06} & 97.13\err{0.13} & 95.71\err{0.17} & 93.85\err{0.37} & 96.41\err{0.01} \\
\midrule
\multicolumn{7}{l}{\textit{Zero-Shot and Augmented Slot F1}} \\
\midrule
mBERT & N/A & 74.98 & 82.61 & 62.27 & 35.75 & 74.05 & 75.71 & 31.21 & 23.75 & 57.54 \\
mBERT + fastalign & N/A & 79.18 & 87.21 & 81.82 & 79.53 & 78.26 & 70.18 & 69.42 & 23.61 & 71.15 \\
mBERT + softalign & N/A & 76.42 & \textbf{89.00} & 83.25 & 79.10 & 76.30 & 79.64 & 78.56 & 61.70 & 78.00 \\
mBERT + TMP & N/A & 83.98 & 87.54 & 85.05 & 82.60 & 81.73 & 79.80 & 77.24 & 44.80 & 77.84 \\
\hdashline
mT5-xxl & 96.19\err{0.19} & 84.60\err{1.20} & 77.03\err{0.59} & 81.00\err{1.31} & 59.29\err{3.76} & 81.62\err{1.06} & 81.72\err{1.20} & 66.28\err{5.12} & 50.50\err{3.37} & 72.76\err{1.25} \\
mT5-xxl + TaF & 95.35\err{0.17} & \textbf{88.26}\err{0.05} & 86.78\err{0.10} & \textbf{87.49}\err{0.41} & \textbf{88.66}\err{0.43} & \textbf{87.30}\err{0.37} & \textbf{86.19}\err{0.25} & \textbf{88.06}\err{0.08} & \textbf{84.47}\err{0.27} & \textbf{87.15}\err{0.14} \\
mT5-xxl + TAP & 95.77\err{0.18} & 85.40\err{0.13} & 84.25\err{0.19} & 81.65\err{0.21} & 82.05\err{0.24} & 82.85\err{0.70} & 84.48\err{0.57} & 86.11\err{0.21} & 82.05\err{1.05} & 83.61\err{0.27} \\
\bottomrule
\end{tabular}
\caption{Intent Accuracy and Slot F1 of our mT5 models on MultiAtis++. Multilingual BERT (mBERT) results are from \newcite{xu-etal-2020-end}. In bold, the best models in the data augmentation scenario.} 
\label{tab:matispp}
\end{center}
\end{table*}

\begin{table*}
\begin{center}
\scriptsize
\setlength{\tabcolsep}{1.1em}
\renewcommand{\arraystretch}{1.2}
\begin{tabular}[b]{p{4cm}cccccccccc}
\toprule
\multirow{1}{*}{\bf MultiAtis++} & \bf en & \bf es & \bf de & \bf zh & \bf ja & \bf pt & \bf fr & \bf hi & \bf tr & \bf Avg(8 langs) \\
\midrule
\multicolumn{7}{l}{\textit{Intent Accuracy}} \\
\midrule
mBERT, Zero-shot & 96.53 & 82.31 & 86.9 & 85.89 & 81.08 & 84.43 & 92.72 & 75.59 & 71.61 & 82.57\\
mBERT, TAP & 97.12 & 95.00 & 96.15 & 93.92 & 91.68 & 96.38 & 96.15 & 94.55 & 79.67 & 92.94\\
mBERT, TaF & 97.16 & 93.32 & 96.60 & 95.29 & 93.58 & 95.19 & 95.67 & 95.11 & 93.48 & 94.78\\
mBERT, Gold & 96.75 & 94.4 & 96.53 & 93.17 & 94.29 & 95.97 & 97.31 & 92.95 & 90.63 & 94.41\\
mBERT, Gold (es lowercased) & 96.75 & 93.73 & 95.41 & 90.48 & 91.38 & 95.97 & 96.53 & 92.61 & 90.63 & 93.34\\
\midrule
\multicolumn{7}{l}{\textit{Slot F1}} \\
\midrule
mBERT, Zero-shot & 95.65 & 43.83 & 31.25 & 67.2 & 50.8 & 48.71 & 45.32 & 40.36 & 29.74 & 44.65\\
mBERT, TAP & 95.79 & 77.48 & 76.05 & 78.69 & 70.25 & 79.38 & 77.89 & 79.36 & 60.24 & 74.92\\
mBERT, TaF & 95.78 & 81.18 & 81.80 & 84.11 & 86.97 & 82.14 & 79.21 & 86.13 & 84.99 & 83.31\\
mBERT, Gold & 95.91 & 72.41 & 90.61 & 90.76 & 89.91 & 87.03 & 87.29 & 86.49 & 85.65 & 86.27\\
mBERT, Gold (es lowercased) & 96.11 & 80.63 & 91.22 & 89.96 & 88.53 & 88.25 & 87.9 & 86.98 & 85.05 & 87.32\\
\bottomrule
\end{tabular}
\caption{Intent Accuracy and Slot F1 of our multilingual BERT (mBERT) model on MultiAtis++.}
\label{tab:mbert}
\end{center}
\end{table*}

\begin{table*}
\begin{center}
\scriptsize
\setlength{\tabcolsep}{1.2em}
\renewcommand{\arraystretch}{1.0}
\begin{tabular}{p{6cm}cc}
\toprule
{\bf Multilingual ATIS} & \bf hi & \bf tr \\
\midrule
\multicolumn{1}{l}{\textit{Multilingual models (trained on all data from all languages)}} \\
\midrule
XLM-R & 62.3 / 85.9 / 87.8 & 65.7 / 92.7 / 86.5 \\
mT5-xxl & 73.01\err{0.30} / 95.04\err{0.06} / 88.93\err{0.09} & 70.68\err{0.63} / 94.13\err{0.37} / 87.69\err{0.26} \\
\midrule
\multicolumn{1}{l}{\textit{Zero-shot models (trained on English only)}} \\
\midrule
XLM-R & 40.3 / 80.2 / 76.2 & 15.7 / 78 / 51.8 \\
mT5-xxl & 40.87\err{8.91} / 91.41\err{0.28} / 68.69\err{7.47} & 14.78\err{2.18} / 84.99\err{0.53} / 51.29\err{3.31} \\
\midrule
\multicolumn{1}{l}{\textit{Augmented data models}} \\
\midrule
XLM-R + translate align & 53.2 / 85.3 / 84.2 & 49.7 / 91.3 / 80.2 \\
mT5-xxl + TaF & \textbf{65.29}\err{0.22} / \textbf{96.23}\err{0.17} / \textbf{84.85}\err{0.09} & \textbf{67.41}\err{0.92} / \textbf{95.15}\err{0.21} / \textbf{85.30}\err{0.18} \\
mT5-xxl + TAP & 63.94\err{0.30} / 96.04\err{0.50} / 84.00\err{0.39}&
58.41\err{0.91} / 95.10\err{0.14} / 82.40\err{0.46}\\
\bottomrule
\end{tabular}
\caption{Results of our mT5 models on Multilingual ATIS. Metrics are Exact Match (EM) accuracy / Intent Accuracy / Slot F1 respectively. XLM-R results are from \newcite{li-etal-2021-mtop}.
\label{tab:multiatis}}
\vspace{-1em}
\end{center}
\end{table*}

\textbf{MTOP}. Table \ref{tab:mtop} contains the results on MTOP. \textit{XLM-R} from \newcite{li-etal-2021-mtop} is a seq2seq model that uses XLM-R as encoder and it is extended with a pointer network. This and the \textit{mt5-xxl} model have a comparable average EM accuracy when trained multilingually with all the available gold data, although \textit{mT5-xxl} has more parameters. In the zero-shot setting, \textit{mT5-large} lags behind \textit{XLM-R} by 7.5 EM points, while \textit{mT5-xxl} already improves over \textit{XLM-R} by 16.3 EM points. When \textit{+TaF} synthetic data is added, \textit{mT5-large+TaF} reaches \textit{XLM-R+TAP}, and \textit{mT5-xxl+TaF} surpasses it by 2.5 points,
indicating that TAF is effective for i18n over strong and weak base models. While we could not run \newcite{li-etal-2021-mtop} model on our data, we can see that \textit{mT5-large+TaF} is able to close all the gap with \textit{XLM-R+TAP}, despite starting from a much lower zero-shot accuracy. \textit{mT5-xxl+TaF} is only 8.6 points behind \textit{mT5-xxl} trained on all the available gold multilingual data and covers 60\% of the gap between zero-shot and full multilingual supervision. \textit{mT5-xxl+TaF} shows a remarkable improvement on German w.r.t. \textit{XLM-R+TAP} and \textit{mT5-xxl+TAP}, probably due to alignment errors caused by the heavy compounding nature of German, as  \newcite{li-etal-2021-mtop} report in their paper too. In the \textit{mT5-xxl+TaF, untokenized} experiment, we do not tokenize the translations for the filler. The results do not significantly change, suggesting that our approach is robust to tokenization and therefore tokenizers and aligners are not necessary. 

Table 3 contains the results on MTOP when training \textit{mt5-xxl} with English gold data and synthetic data generated by TAP for all the other languages. Out-of-the-box TAP is well behind zero-shot. 
With POS-based postprocessing, we see a significant improvement in all languages. Except for Thai, all languages are well above zero-shot performance. This shows that human-engineering is essential for TAP to perform well. Note that the preposition and determiner exclusion rule is not being applied for the \texttt{DATE\_TIME} slots, according to the labeling trend we have observed in the MTOP dataset. On the other hand, the filler is able to learn this trend by itself and no heuristics are needed. The experiment where we keep only the whitespace-segmented synthetic data reaches the best performance with a significant bump in Thai, but it is still \texttildelow4 EM points below that of the filler on average. This shows that high-quality alignments are paramount for TAP to work well. The filler completely eliminates the need of an aligner and achieves better results. Note that for the other tables, we only included the results from the best TAP configuration.
\\\\
\textbf{MultiATIS++}. In Table \ref{tab:matispp}, we compare our approach with the mBERT models from \newcite{xu-etal-2020-end} that use synthetic data obtained by projecting labels with \textit{fastalign} alignments \cite{dyer-etal-2013-simple}, attention weights (\textit{softalign}) and \textit{TMP} linguistic features \cite{Jain2019}. mT5-xxl has remarkable zero-shot Intent Accuracy and Slot F1, even without synthetic data. With the latter, the average Slot F1 is \texttildelow10 points higher than the best mBERT baselines and the result variance decreases significantly w.r.t. zero-shot. Data produced with fastalign degrades performance for French and Turkish, while TaF synthetic data always leads to better results and contributes to set SOTA performance in data augmentation settings for mT5. We achieve a 52.83\% Relative Reduction in Error (RRIE) in Slot F1 w.r.t. zero-shot, and compared to using all gold data, we close the full gap in Intent Accuracy and reduce the difference in Slot F1 to only \texttildelow4.5 points. TaF consistently outperforms TAP in all languages, with more pronounced differences in Slot F1.

In Table \ref{tab:mbert} we report results on MultiAtis++ with an mBERT \cite{Devlin2019} based parser, in order to understand how effective our synthetic data method is with models with less parameters and lower complexity than mT5.
We use the cased mBERT encoder to obtain word representations: a linear layer on top of the \texttt{[CLS]} token performs intent classification, while a linear layer on top of the first wordpiece of each sentence token is used for slot tagging. Similarly to what we have observed in the mT5-xxl experiments, our synthetic data closes the full gap (between zero-shot and gold) in average Intent Accuracy, while the gap in average Slot F1 is reduced to less than 3 points. This shows the effectiveness of TaF for models with different power and capacity. It is worth noting that in the multilingual gold setting Spanish was underperforming TaF. The reason seemed to be that Spanish training data is mixcased while test data is lowercased. If we lowercase Spanish training data, we see a significant improvement in Spanish Slot F1. Note that the zero-shot and gold performance of our mBERT model is below that of the implementation in \newcite{xu-etal-2020-end}. We suspect this is due mBERT model differences: our model has about 110M parameters
while \newcite{xu-etal-2020-end} report more than 166M parameters. Despite the lower zero-shot performance, our mBERT model with TaF is more than 5 points better in average Slot F1 across 8 languages than the best data augmentation method from \newcite{xu-etal-2020-end}.
\\\\
\textbf{Multilingual ATIS}. Table \ref{tab:multiatis} confirms the MultiAtis++ results also for this dataset. mT5-xxl is more effective than XLM-R at zero-shot intent classification but not at Slot F1. With the help of our synthetic data, \textit{mT5-xxl+TaF} reaches SOTA performance on the task, recovering 51.6\% and 69.8\% of the full supervision gap in Slot F1 w.r.t zero-shot, on the Hindi and Turkish evaluation sets respectively. We observe TaF outperforms TAP on EM, particularly on highly agglutinative Turkish.
\\\\
\textbf{General Remarks}. The relative deltas in performance across datasets on same languages may be explained by the heterogeneous domains and by the annotation structure. In addition, the starting pretrained models have different quality across languages as shown in  “Zero-shot models” in our tables and as also noted in \newcite{Conneau20XLMR} (e.g., XLM-R performs particularly well on low-resource languages). Pretraining quality typically transfers to fine-tuned models.

\begin{table}
\scriptsize
\centering
\renewcommand{\arraystretch}{0.95}
\begin{tabular}{crr}
\toprule
{\bf Language } & Filler Errors & Zero-shot Parser Errors \\
\midrule
{de} & {459 (2.93\%)} & {3607 (23.02\%)} \\
{es} & {97   (0.62\%)} & {2977 (19.00\%)} \\
{fr} & {98   (0.63\%)} & {3001 (19.15\%)} \\
{hi} & {241  (1.53\%)} & {4811 (30.71\%)} \\
{th} & {1369 (8.74\%)} & {5556 (35.46\%)} \\
\midrule
{Total} & {2264 (2.89\%)} & {19952 (25.47\%)} \\
\bottomrule
\end{tabular}
\caption{Number and \% of instances with errors that are matched by our heuristic filters.}
\label{tab:filler-errors}
\vspace{-1em}
\end{table}

\begin{table*}
\scriptsize
\centering
\renewcommand{\arraystretch}{0.95}
\begin{tabular}{cll}
\toprule
Lang & Utterance & Representation \\
\midrule
 & (1) \textit{Hallucination of pronouns} & \\
\midrule
en & What reminders do \textbf{we} have this weekend ? & [IN: ... [SL:PERSON\_REMINDED \textbf{we} ] [SL:DATE this weekend ] ] \\
es & Qué recordatorios hacemos este fin de semana ? & [IN: ... [SL:PERSON\_REMINDED \textbf{nosotros} ] [SL:DATE este fin de semana ] ] \\
\midrule
 & (2) \textit{Confusion around prepositions and determiners} & \\
\midrule
en & cancel reminder to call dentist & [IN: ... [SL:TODO [IN:CREATE\_CALL [SL:CONTACT dentist ] ] ] ] \\
es & cancelar recordatorio para llamar \textbf{al} dentista & [IN: ... [SL:TODO [IN:CREATE\_CALL [SL:CONTACT \textbf{el} dentista ] ] ] ] \\
\midrule
 & (3) \textit{Slot words reordering} & \\
\midrule
en & What year did T. Woods \textbf{turn pro} ? & [IN: ... [SL:CONTACT T. Woods ] [SL:EVENT \textbf{turn pro} ] ] \\
es & En qué año \textbf{se convirtió }T. Woods \textbf{en profesional} ?
 & [IN: ... [SL:CONTACT T. Woods ] [SL:EVENT \textbf{se convirtió en profesional} ] ] \\
\midrule
 & (4) \textit{Hallucination of unaligned/missing words} & \\
\midrule
en & Play some rap \textbf{music} & [IN: ... [SL:MUSIC\_GENRE rap ] [SL:MUSIC\_TYPE \textbf{music} ] ] \\
es & toca algo de rap & [IN: ... [SL:MUSIC\_GENRE rap ] [SL:MUSIC\_TYPE \textbf{music} ] ] \\
\midrule
en & Will it be hot \textbf{out} today ? & [IN: ... [SL:WEATHER\_ATTR hot ] [SL:LOC \textbf{out} ] [SL:TIME today ] ] \\
es & Va a hacer calor hoy ? & [IN: ... [SL:WEATHER\_ATTR calor ] [SL:LOC \textbf{Alicante} ] [SL:TIME hoy ] ] \\
\midrule
 & (5) \textit{Compound or word splitting} & \\
\midrule
en & Delete the \textbf{homework} reminder & [IN: ... [SL:TODO \textbf{homework} ] ] \\
de & Löschen Sie die \textbf{Hausaufgaben}erinnerung & [IN: ... [SL:TODO \textbf{Hausaufgaben} ] ] \\
\bottomrule
\end{tabular}
\caption{Examples where our filler generates spans that cannot be found in the input translation (in Spanish or German). \textit{en} rows contain the original English utterance and parse. Intents are omitted and some slots are shortened for readability.}
\label{tab:decoupled-comparison}
\end{table*}

\section{Analysis of the Filler Output}

In this section, we analyze the output of our filler trained on the English MTOP data and run on translated MTOP.
We use two simple heuristic filters to understand how good the filler is at reproducing the signature provided in the input and how much it suffers from hallucination. Therefore we count (i) how many times the input and output signatures differ (ignoring slot orders); (ii) for how many utterances the output slots contain word spans which cannot be found in the input utterance.

Table \ref{tab:filler-errors} contains the number of examples (and the \%s for each language) triggering our filters. The last row summarizes the numbers for the total of about 75k utterances (\texttildelow15k English MTOP training instances translated to 5 languages). In addition to the filler statistics, we compute the same numbers for a model that does not have access to the parse signature, i.e., a zero-shot parser trained on English. As we can see, the outputs of the filler contain errors in only 2.89\% of cases. Of these, 0.5\% parses are malformed, 3.7\% have mistakes in the signatures and 96\% have hallucination errors. We can conclude that the filler is able to reproduce input signatures and the only issues are due to wrong tokens put in the slots. On the contrary, the 25\% outputs with mistakes from the zero-shot parser are dominated by signature mistakes, which are 76\% of the total. Hallucination errors amount to 28\%.

Table \ref{tab:decoupled-comparison} contains interesting examples matched by our heuristic filters. Hallucinations may happen when some words are dropped in the translation. In (1), the pronoun is dropped and the model generates the relevant first person plural pronoun in Spanish. In (4), the word ``music'' is not contained in the translation but still relevant, while ``Alicante'' is a quite random choice for the location slot. Other frequent issues are related to the choice of prepositions and determiners as in (2), where the latter is often preferred by the model. Example (3) is an interesting case of word reordering that highlights a well known issue in the i18n of span labeling annotations, namely span splitting.
The generative filler is able to reorder the phrase back. Finally, we highlight example (5). In German, a compound rich language, the noun ``homework'' forms a compound with the noun ``reminder''. The filler is able to split the compound noun, thanks to its subword output vocabulary, and put the relevant part in the \texttt{TODO} slot. How useful this is ultimately depends on the annotation guidelines defined for the i18n languages (e.g., allowing and supporting subword annotations).
The relatively low number of errors and their nature explain why we are able to use all the synthetic data produced by our method to train the final parsers. We experimented by filtering out synthetic examples with the aforementioned heuristics but we did not register any improvement on the final performance.

\section{Conclusions and Future Work}

In this paper, we proposed a novel Translate-and-Fill synthetic data generation approach which requires less engineering effort than TAP. TaF leverages NMT, multilingual pretrained seq2seq models and task labels, at the same time removing the need of aligners and tokenizers. Our filler model, trained on English data only, works remarkably well on other languages and enables improvements on multiple semantic parsing datasets in synthetic data scenarios.
As future work, we plan to explore applications of the filler to (i) other i18n synthetic data generation tasks that require span alignment and to (ii) in-language data augmentation, e.g., using paraphrases to improve parsing accuracy of intent and slots with little annotated data.

\section*{Acknowledgments}
We would like to thank Melvin Johnson and the anonymous reviewers for their constructive feedback, useful comments and suggestions.

\section*{Ethical Considerations}

This work does not use sensitive data, or language models for uncontrolled generation. The output of the parser is not user facing, therefore engineers can easily intervene to eliminate potentially harmful hallucinations of the seq2seq model by simple filtering. One concern could be the energy consumption of the experiments and to mitigate that, we did not perform hyper-parameter tuning and limited the experiment reruns.  
 
\bibliography{anthology}
\bibliographystyle{acl_natbib}

\end{document}